# LSTM-BASED TEXT GENERATION: A STUDY ON HISTORICAL DATASETS

**Mustafa Abbas Hussein Hussein[1*], Serkan Savaş[2]**

[*1]Çankırı Karatekin University, Graduate School of Natural and Applied Sciences, Department of Electronics and Computer Engineering, Çankırı, Türkiye. ORCID Code: 0009-0002-9219-5695

[2]Kırıkkale University, Faculty of Engineering and Natural Sciences, Department of Computer Engineering, Kırıkkale, Türkiye. ORCID Code: 0000-0003-3440-6271

## ABSTRACT

This paper presents an exploration of Long Short-Term Memory (LSTM) networks in the realm of text generation, focusing on the utilization of historical datasets for Shakespeare and Nietzsche. LSTMs, known for their effectiveness in handling sequential data, are applied here to model complex language patterns and structures inherent in historical texts. The study demonstrates that LSTM-based models, when trained on historical datasets, can not only generate text that is linguistically rich and contextually relevant but also provide insights into the evolution of language patterns over time. The finding presents models that are highly accurate and efficient in predicting text from Nietzsche's works, with low loss values and a training time of 100 iterations. The accuracy of the model is 0.9521, indicating high accuracy. The model's loss is 0.2518, indicating its effectiveness. The accuracy of the model in predicting text from Shakespeare's works is 0.9125, indicating a low error rate. The model's training time is 100, mirroring the efficiency of the Nietzsche dataset. This efficiency demonstrates the effectiveness of the model's design and training methodology, especially when handling complex literary texts. This research contributes to the field of natural language processing by showcasing the versatility of LSTM networks in text generation and offering a pathway for future explorations in historical linguistics and beyond.

## INTRODUCTION

Natural language generation (NLG) is a software method that produces output in natural language, with text generation (TG) being one of its applications. TG, as defined, is a subfield of artificial intelligence and computational linguistics that focuses on creating computer systems capable of producing understandable texts in human languages from non-linguistic information representations. There's some debate about whether TG inputs must be non-linguistic, but the consensus is that text is the outcome of any TG process. TG is used in various applications, such as generating reports, image captioning, and chatbots, and is similar to human verbal or written expression. In recent years, deep learning algorithms have produced successful results in many different fields. Deep learning algorithms such as convolutional neural networks, recurrent neural networks, and deep belief nets are used in various fields such as health, industry, agriculture, education, security, etc (Ayan, 2022; Bütüner & Calp, 2022; Buyrukoğlu et al., 2021; Güler & Polat, 2022). Recurrent Neural Networks (RNNs) are a class of artificial neural networks where connections between nodes can form loops, allowing for temporal dynamic behavior. Built from feedforward neural networks, RNNs use their internal state (memory) to process variable-length input sequences, making them suitable for tasks like speech and handwriting recognition. RNNs can theoretically process any input sequence, distinguishing them from convolutional neural networks. While RNNs are infinite impulse response networks, finite impulse recurrent networks, which are directed acyclic graphs, can be unrolled into strictly feed forward networks, unlike RNNs, which are directed cyclic graphs and cannot be unrolled. (Pawade et al., 2018) aimed to create new stories from a compilation of existing ones, exploring two methods: one using varied themes and fonts, and another using interconnected narratives and similar characters. The system's output was evaluated for grammar accuracy, event causality, curiosity level, and originality. (Khatri et al., 2015) developed a deep learning and natural language processing (NLP) based framework for generating product context on e-commerce websites, involving five unsupervised steps such as keyword collection, duplication prevention, sentence selection, sentence creation using RNN and LSTM, and contextual organization with Text Rank. (Thomaidou et al., 2013) proposed a method for generating persuasive text from advertisements, using steps like information extraction and sentimental evaluation. (Jain et al., 2017) described a method for creating a story from brief narrations using phrase-based statistical machine translation to create phrase-maps for the target language. (Li et al., 2013) introduced "SCHEHERAZADE," a story creation method using crowd sourced plot graphs with standard, conditional, and optional events linked by mutual exclusion and precedence constraints.

(Nallapati et al., 2016) created an encoder-decoder RNN with a large vocabulary trick and attention for abstract text summarization, also using a switching decoder/pointer architecture and hierarchical attention.

(Jaya & Uma, 2010) presented an ontology-based system for story creation, where ontology provides attributes to selected parameters, maintaining the domain's context. (Jaya & Uma, 2011) enhanced the Proppian's system with semantic reasoning and ontology for a common conception, improving its capabilities. (Le & Le, 2013) proposed an abstract text synthesis method using discourse rules and syntactic constraints, followed by a Word graph for coherent sentence generation. (Sakhare & Kumar, 2014) suggested a hybrid text summarization approach combining sentence structure and attribute-based algorithms, involving preprocessing, neural network training, and feature score calculation. (Kavila & Radhika, 2015) developed an extractive text summarization technique for IEEE papers, focusing on modified sentence symmetry and weighting techniques. (Hatipoglu & Omurca, 2016) built a smartphone app for summarizing Turkish Wikipedia data, enhancing a hybrid technique with structural and semantic features based on Wikipedia and LAS. (Colon et al., 2014) mentioned a professional writing prompts application using random WordNet words, ConceptNet, and concept correlation for plot development and evaluation. (Mehta et al., 2016) created a system using the Generate and Rank approach and NLG, with inputs like theme and duration, employing Centric Theory, action graph, local coherence, and a lexical database for content identification, followed by planning and rule-based story generation.

In this study, we aim to improve NLP by developing a refined language model using LSTM layers. The model focuses on contextual understanding for enhanced text generation, comparing its performance against historical datasets like Nietzsche and Shakespeare. The goal is to surpass traditional text generation methodologies by integrating advanced contextual analysis into language model training. The study uses a state-of-the-art RNN architecture with LSTM layers to test this hypothesis.

The remaining part of this study; methodology provided in Section 2. Results and discussion in Section 3, which evaluates and contrasts current models, and the conclusion section summarizes the findings and suggests future research directions.

**MATERIAL AND METHOD**

This research utilizes two datasets pertaining to historical novels: those of Shakespeare (LiamLarsen, 2017) and Nietzsche (Kris, 2018). The study uses these two datasets to evaluate model performance. The first dataset, which includes all of Shakespeare's plays, is organized for text analysis or NLP tasks. It provides detailed location information for each line within the play's structure, identifies the specific Shakespearean play, and names the character speaking the line. The Line Being Spoken column contains the actual words written by Shakespeare and spoken by the characters in his plays, essential for textual analysis. This dataset can be used for various purposes, such as studying Shakespeare's writing style, performing textual analysis to find patterns or themes across different plays, training machine learning models for tasks like text generation or sentiment analysis (LiamLarsen, 2017). The second dataset, the Nietzsche Texts dataset, is a comprehensive collection of English text specifically designed for natural language processing and text generation tasks. It is recommended for educational and experimental use, as LSTM networks are well-suited for learning sequences and patterns in text (Kris, 2018).

After preprocessing to remove citations and numbering, these datasets were divided into training and testing sets with a proportion of 70% and 30%, respectively. To test the efficacy of our method, we constructed an LSTM model in this study. To process text data, computers must convert it into a readable and understandable format. This conversion can be achieved through techniques such as one-hot encoding or word embedding, which create a dictionary of words recurring in the text. These models, simple to train, retain essential information for precise predictions. In this approach, the input is paired with the output, and the RNNs are implemented using LSTM layers via the Keras library. Our model employs a 3-layer architecture, where the first layer is responsible for information storage and the output layer for character prediction. To interpret the output, a dictionary is used to convert characters, which were initially represented as numbers, back into their original letter form. Figure 1 shows the process that used in this methodology.

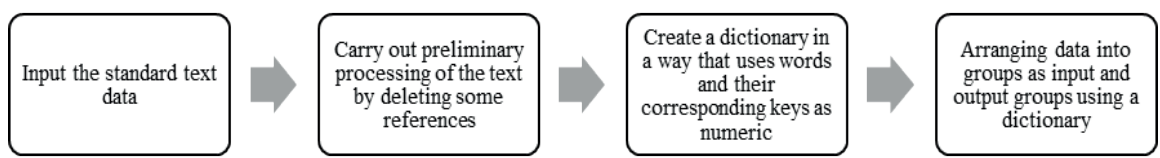

**Figure 1. Flow chart of dataset and preprocessing**

The reason of using the recurrent LSTM network is that it has a structure that is suitable for sequenced sentence vectors (Bilen & Horasan, 2022). The foundation of our proposed model lies in the use of LSTM layers. Instead of merely predicting the next character based on identified patterns and outputs, our model is capable of discerning the appropriate context for generating the desired text. This model builds upon extensive research in RNNs) renowned for their ability to retain short-term memory and their application in complex tasks including NLP. Also, Figure 2 depicts the process in a flowchart.

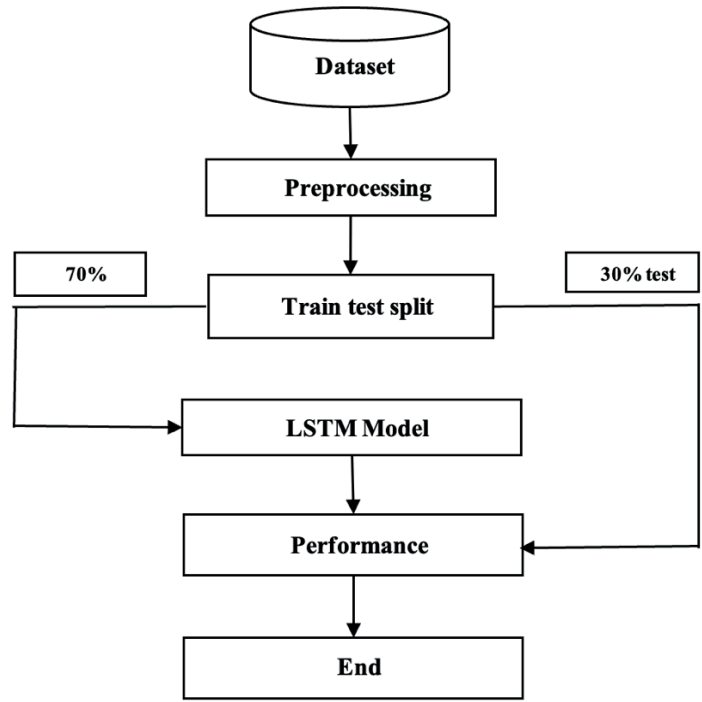

**Figure 2. Flowchart of the proposed study**

When training RNNs, we can effectively sequence words and sentences consecutively, albeit without considering the surrounding context. Transitioning from machine translation to a question-and-answer system, we employ a network architecture of a similar nature. To obtain grammatically accurate but potentially illogical statements from input strings, it is necessary to use seq2seq encoding and decoding networks. The training and word generation process of the model might result in sentences that lack cohesion. However, through comprehensive training of the proposed model using all its units collectively, we aim to accurately identify the appropriate phrases and effectively reduce lexical ambiguity.

Consequently, we proposed a model that utilizes RNNs to generate text in alignment with the input data and within the relevant context. The model is trained using input data, output data, and word relationships, all within the appropriate context. This is achieved through a language model, enabling the network to generate sentences based on a defined set of contextual terms. The model's evaluation is based on several criteria: it aims to avoid phrase repetition, generate grammatically coherent new sentences, and ensure that the output text aligns with the input context in terms of linguistic significance, regardless of sentence quality. Figure 3 shows the model structure with its parameters.

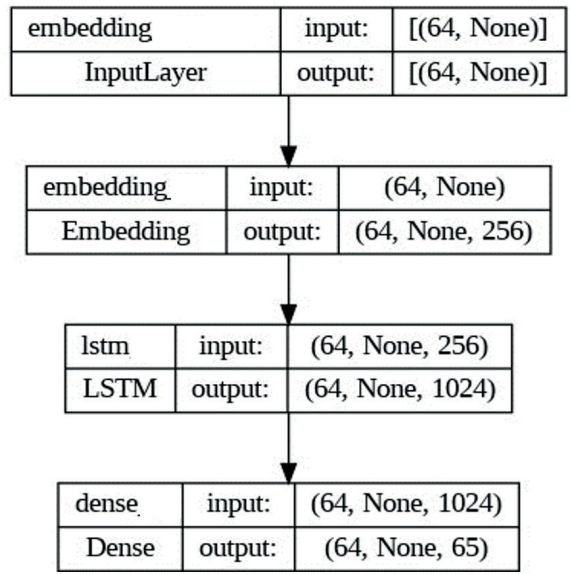

**Figure 3. Model parameters with layer types and output shape**

## RESULTS AND DISCUSSION

In this section, the performance of proposed model have been evaluated. According to Nietzsch data, Figure 4 shows how the model's accuracy and loss improve as it learns from the training data over multiple epochs. It starts with a relatively low accuracy and gradually increases as the model refines its parameters. Also, the model loss decreases over epochs. Initially, the model makes significant errors in its predictions, and these errors gradually reduce as training continues. The goal is to minimize the loss as much as possible. Balancing the loss and accuracy curves is important for ensuring that the model is learning effectively and generalizing well to unseen data.

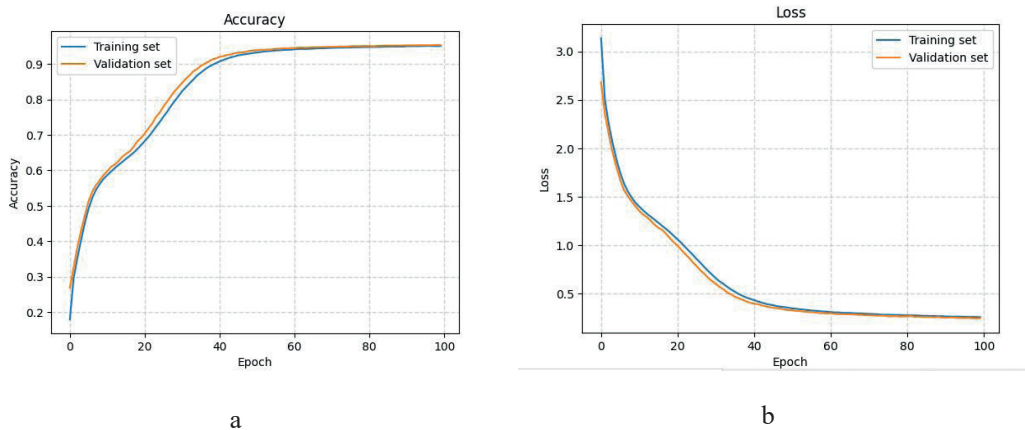

a                                                              b

**Figure 4. Model performance based on Nietzsche dataset a) accuracy b) loss**

Additionally, when the model was applied to another dataset, it demonstrated robust performance in text generation, characterized by a high level of accuracy and low losses. Figure 5 illustrates the model's performance based on the Shakespeare dataset.

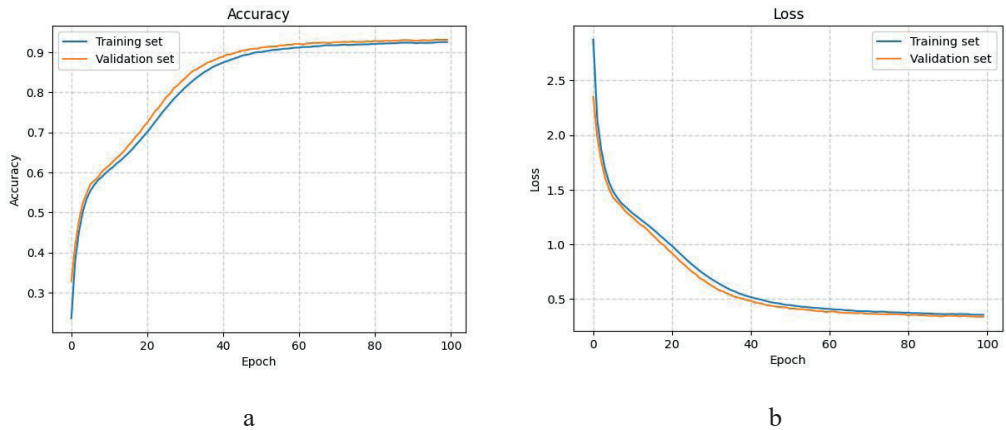

**Figure 5. Model performance based on Shakespeare dataset a) accuracy b) losses**

The results gathered were analyzed and compared with the findings from previous studies conducted on Shakespeare and Nietzsche dataset.

Table 1 compares the performance of models trained on two datasets: one from Nietzsche's works and another from Shakespeare's works. The performance metrics include accuracy, loss, and the number of iterations used in the training process. (Gao & Glowacka, 2016) reported an accuracy of 0.758 on the Nietzsche dataset, requiring 160 iterations. The current study developed a model with a higher accuracy of 0.9521, with a loss of 0.2518, over 100 iterations. (Hussain et al., 2021) reported an accuracy of 0.746 on Shakespeare's dataset, with a loss of 1.020 over 500 iterations. The current study's model on Shakespeare's dataset showed an improved accuracy of 0.9125, with a lower loss of 0.3876, achieved in 100 iterations. The current study outperformed previous studies in terms of accuracy and loss metrics, indicating more efficient learning. The loss value is not provided for Gao's model on the Nietzsche dataset. So, the table highlights the effectiveness and efficiency of the models developed in this study, with significant improvements in accuracy and loss within a reduced number of training iterations.

**Table 1. Compares the performance of models trained on two datasets with other studies**

|  | MODELS | ACCURECY | LOSS | NUMBER ITERATIONS |
|---|---|---|---|---|
| **NIETZSCHE DATASET** | (Gao & Glowacka, 2016) | 0.758 | - | 160 |
|  | This study | 0.9521 | 0.2518 | 100 |
| **SHAKESPEARE DATASET** | (Hussain et al., 2021) | 0.746 | 1.020 | 500 |
|  | This study | 0.9125 | 0.3876 | 100 |

## CONCLUSION

This study demonstrates the superior performance of LSTM in generating text from Shakespearean and Nietzsche's dataset. The efficiency of the model is attributed to its low training requirement of 100 iterations, which is a significant reduction compared to other models. The contextual training approach, employing LSTM layers to capture long-term dependencies, enables the extraction of meaningful context vectors and the clustering of words based on semantic proximity. This methodology stands in contrast to previous models, which reported lower accuracies and required more iterations for training. LSTM also outperforms others in terms of accuracy and loss, indicating a more robust performance in minimizing errors. The reduced number of iterations required for optimal performance further underscores its effectiveness in text generation tasks. The findings of this study highlight the significance of innovative training techniques and contextual understanding in achieving high-quality text generation, thereby setting a new benchmark for future research in the field.